\documentclass{article}
\usepackage{hyperref}       % hyperlinks
\usepackage{url}            % simple URL typesetting
\usepackage{amsfonts}       % blackboard math symbols
\usepackage{algorithm}
\usepackage{algorithmic}
\usepackage{braket}
\usepackage{fullpage}
\usepackage[colorinlistoftodos]{todonotes}

\usepackage{amsmath,amsfonts,amssymb}
\usepackage{mathtools}
\usepackage{amsthm} % better to load after ams math
\usepackage{latexsym}
\usepackage{relsize}

\usepackage[capitalise]{cleveref}
\usepackage{xcolor}
\usepackage{dsfont}

\newcommand{\CH}{\text{CH}}
\newcommand{\F}{\mathbb{F}}

\def\H{{\mathcal H}}

\def\W{{\mathcal W}}
\def\R{{\mathcal R}}

\newcommand{\ftil}{\tilde{f}}

\newcommand{\D}{\mathcal{D}}

\def\W{{\mathcal W}}
\def\mA{{\mathcal A}}
\def\fhat{{\hat{f}}}

\newcommand{\bzero}{\ensuremath{\mathbf 0}}
\newcommand{\ba}{\ensuremath{\mathbf a}}
\newcommand{\y}{\ensuremath{\mathbf y}}
\newcommand{\z}{\ensuremath{\mathbf z}}

\newcommand{\K}{\ensuremath{\mathcal K}}

\renewcommand\F{\mathcal{F}}

\def\C{{\mathcal C}}

\def\x{\mathbf{x}}
\def\y{\mathbf{y}}

\def\regret{\mbox{{Regret}}}

\newcommand{\ignore}[1]{}

\newcommand{\eh}[1]{\noindent{\textcolor{magenta}{\{{\bf EH:} \em #1\}}}}

\theoremstyle{plain}
\newtheorem{theorem}{Theorem}
\newtheorem{lemma}[theorem]{Lemma}
\newtheorem{corollary}[theorem]{Corollary}

\newtheorem{assumption}{Assumption}

\newtheorem*{theorem*}{Theorem}
\newtheorem*{lemma*}{Lemma}
\newtheorem*{corollary*}{Corollary}
\newtheorem*{proposition*}{Proposition}
\newtheorem*{claim*}{Claim}
\newtheorem*{fact*}{Fact}
\newtheorem*{observation*}{Observation}
\newtheorem*{assumption*}{Assumption}

\theoremstyle{definition}
\newtheorem{definition}[theorem]{Definition}

\newtheorem*{definition*}{Definition}
\newtheorem*{remark*}{Remark}
\newtheorem*{example*}{Example}

 \theoremstyle{plain}
\newtheorem*{theoremaux}{\theoremauxref}
\gdef\theoremauxref{1}

\DeclareMathAlphabet{\mathbfsf}{\encodingdefault}{\sfdefault}{bx}{n}

\DeclareMathOperator*{\argmin}{arg\,min}

\newcommand{\mythreecases}[6] {{
\left\{
\begin{array}{ll}
    {#1} & {\;\text{#2}} \\[1ex]
    {#3} & {\;\text{#4}} \\[1ex]
    {#5} & {\;\text{#6}}
\end{array}
\right. }}

\renewcommand{\O}{O}

\newcommand{\E}{\mathbb{E}}

\newcommand{\reals}{\mathbb{R}}
\newcommand{\eps}{\varepsilon}

\renewcommand{\leq}{~\le~}

\let\oldtfrac\tfrac
\renewcommand{\tfrac}[2]{\smash{\oldtfrac{#1}{#2}}}

\def\dist{{\bf Dist}}

\let\nablaold\nabla
\renewcommand{\nabla}{\nablaold\mkern-2.5mu}

\title{Boosting for Online Convex Optimization}

\author{%
Elad Hazan $^{1\,2}$ \qquad \qquad Karan Singh $^2$\\
$^1$ Google AI, Princeton\\
$^2$ Computer Science, Princeton University\\
\texttt{\{ehazan, karans\}@princeton.edu}
}

\begin{document}

\maketitle

\begin{abstract}
We consider the decision-making framework of online convex optimization with a very large number of experts. This setting is ubiquitous in contextual and reinforcement learning problems, where the size of the policy class renders enumeration and search within the policy class infeasible.

Instead, we consider generalizing the methodology of online boosting. We define a weak learning algorithm as a mechanism that guarantees multiplicatively approximate regret against a base class of experts. In this access model, we give an efficient boosting algorithm that guarantees near-optimal regret against the convex hull of the base class. We consider both full and partial (a.k.a. bandit) information feedback models. We also give an analogous efficient boosting algorithm for the i.i.d. statistical setting. 

Our results simultaneously generalize online boosting and gradient boosting guarantees to contextual learning model, online convex optimization and bandit linear optimization settings. 
\end{abstract}

\section{Introduction}
In the classical problem of prediction from expert advice \cite{cesa1997use}, a learner iteratively makes decisions and receives loss according to an arbitrarily chosen loss function. Learning in terms of guaranteeing an absolute bound on the loss incurred would be a hopeless task, but the learner is assisted by a pool of experts (or hypotheses). The goal of the learner is thus to minimize regret, or the difference of total loss to that of the best expert in hindsight.

For this canonical problem, numerous methods have been devised and refined, some notably based on the multiplicative weights algorithm \cite{LITTLESTONE1994212,arora2012multiplicative}. It is well established that the regret can be bounded by $O(\sqrt{T \log |\mathcal{H}|})$, where $|\mathcal{H}|$ is the number of experts, and that this is tight. However, in many problems of interest, the class of  experts is too large to efficiently manipulate. This is particularly evident in contextual learning, where the experts are {\it policies} -- functions mapping contexts to action. In such instances, even if a regret bound of $O(\sqrt{T \log |\mathcal{H}|})$ is meaningful, the algorithms achieving this bound are computationally inefficient; their running time is linear in $|\mathcal{H}|$.

One approach to deal with this computational difficulty is to grant the leaner a best-in-hindsight (or ERM) oracle. This has been explored in both the stochastic bandit \cite{langford2008epoch,agarwal2014taming,dudik2011efficient} and (transductive) online settings \cite{rakhlin2016bistro,syrgkanis2016improved,syrgkanis2016efficient,hazan2016computational}. 

In this paper, we consider a different approach towards addressing computational intractability, motivated by the observation that it is often possible to design simple {\it rules-of-thumb} \cite{viola2001rapid} that perform slightly better than random guesses. We propose that the learner has access to a ``weak learner" - an computationally cheap mechanism capable of guaranteeing multiplicatively approximate regret against a base hypotheses class. Such considerations arise in the theory of boosting \cite{schapire2012boosting}, which was originally devised in the context of binary classification in the statistical/i.i.d. setting. A weak learner, defined as one which is capable of better-than-random classification, is used by a boosting algorithm to create an accurate classifier. By analogy, the regret-based notion of weak learner we define here is a natural extension of the concept of weak learning to online convex optimization. 

We design efficient algorithms that when provided weak learners compete with the convex hull of the base hypotheses class with near-optimal regret. We further consider different information feedback models: both full gradient feedback, as well as linear bandit (function value) feedback, and derive efficient sublinear regret algorithms for these settings. 

\subsection{Setting and contributions}

%To illustrate the challenge, it is instructive to consider the fundamental problem of prediction from expert advice.  Over $T$ iterations, $|\H|$  experts observe a context and predict a binary outcome, and incur a loss as a function of their prediction. 
%The Weighted Majority algorithm \cite{LITTLESTONE1994212} applies to this problem, and guarantees a regret of $O(\sqrt{T \log |\H|})$, and this is tight. However, if $|\H|$ is extremely large or even infinite, maintaining the weights is prohibitive computationally. Our results apply to this setting with asymptotically the same regret bound and a running time that is independent of $|\H|$. 

The setting of online convex optimization (OCO) generalizes the problem of prediction from expert advice to a general convex decision set $\K \subseteq \reals^d$, and adversarially chosen convex loss functions $f_t : \reals^d \mapsto \reals$. We consider a hypothesis class $\H \subseteq \C \mapsto \K$, where each hypothesis $h\in \H$ maps a context (or feature vector) in the context set $c_t\in \C$ to  an action $h( c_t ) \in \K$. 

%Numerous methods capable of doing so are known, see e.g. comprehensive texts \cite{CbL,hazan2019introduction}. However, most assume that we have enumerate access to the set $\H$, and depend on its cardinality in some way. 
%To avoid this dependence, we consider an alternative to direct access to $\H$. A weak learner for the OCO setting is defined as follows. 

\begin{definition}%[$\gamma$-weak OLO and OCO algrithms]
\label{online_agnostic_wl}
    An online learning algorithm $\W$ is a $\gamma$-\textbf{weak OCO learner (WOCL)} for a class $\H$ and {\it edge} $\gamma \in (0,1)$,
    if for any sequence of contexts chosen from $\C$ and unit\footnote{$f(x)$ is a unit loss function if $\max_{x\in \K} f(x)-\min_{x\in K} f(x) \leq 1$. This scaling is w.l.o.g. and all results hold with a different constant otherwise.} \emph{linear} loss functions $f_1,...,f_T$: 
\begin{align*}
    %\label{eq:wl-stepwise}
  \sum_{t=1}^T f_t(\W(c_t))  \le& \gamma \cdot \underset{h \in \H}{\min} \sum_{t=1}^T f_t(h(c_t)) + (1-\gamma) \sum_{t=1}^T f_t^\mu + \regret_{T}(\W ),
\end{align*}
where $f^\mu = \int_{\x \in \K } f(\x) d\mu$ is the average value of $f$ under the uniform distribution $\mu$.
\end{definition}

In classical boosting, a weak learner is an algorithm that offers an edge over a random guess. Since $f_t^\mu$ is precisely the loss associated with a uniform random guess, the above definition is a generalization of the classic notion. Also, such a definition is invariant under constant offsets to the function value.

It is convenient to henceforth assume that the convex decision set is centered, that is $\int_{\x\in K} \x d\mu =0$. If so, note that for any linear function $f:\R^d\to \R$, $f_t^\mu = f(\int_{\x} \x d\mu)=0$. With this adjustment, we can rephrase the $\gamma$-WOCL learning promise as: 
\begin{equation} \label{eqn:simpleWOLL}
 \sum_{t=1}^T f_t(\W(c_t))  \le \gamma \cdot \underset{h \in \H}{\min} \sum_{t=1}^T f_t(h(c_t))  + \regret_{T}(\W ).
 \end{equation}

Let $\CH(\H)$ be the convex hull of the hypothesis class. Our goal is to design an algorithm $\mA$ which minimizes regret vs. the best hypothesis in the convex hull of the base class.
$$ \regret = \sum_{t=1}^T f_t( \mA(c_t)  )  -  \underset{h \in CH(\H)}{\min} \sum_{t=1}^T f_t(h(c_t))   .
$$

In this model, our main result is an efficient algorithm, whose guarantee is given by:

\begin{theorem}[Main] \label{thm:main}
Algorithm~\ref{alg:ocoboost} with parameters $\gamma,N$, maintains $N$ copies of $\gamma$-WOCL, and generates a sequence actions $\x_1,...,\x_T$, such that 
\[ \boxed{ \regret =  O \left(  \frac{T}{ \gamma\sqrt{ N } }  + \frac{1}{\gamma} {\text{Regret}_{T}(\W)} \right). } \]
\end{theorem}

The formal theorem which we state and prove below has explicit constants that depend on the diameter of $\K$ (as opposed to $|\H|$) and the Lipschitz constants of the loss functions. These are hidden for brevity. 
Notably,  Algorithm \ref{alg:ocoboost} achieves the following goals:
\begin{enumerate}
    \item Its running time does {\bf not} depend on $|\H|$, but rather only on the other natural parameters of the problem, notably the parameter $N$.
    
    \item The regret it guarantees holds with respect to a stronger benchmark than the best single hypothesis in the base class. It competes with the best convex combination of hypotheses from the base class.
    
    \item Finally, the regret guarantee is not multiplicatively approximate anymore; it is directly with respect to the best convex combination without the $\gamma$ factor.
\end{enumerate}
\subsection{Related Work}

Boosting and ensemble methods are fundamental in all of machine learning and data science, see e.g.  \cite{Breiman01} for a retrospective viewpoint and  \cite{Schapire2012} for a comprehensive text.  Originally boosting was studied for statistical learning and the realizable setting. For a framework that encompasses the different boosting methods see \cite{brukhim2020online}. 

More recently boosting was extended to real-valued prediction via the theory of gradient boosting \cite{friedman2002stochastic}, and to the online setting \cite{leistner2009robustness, chenonline, chen2014boosting, beygelzimer2015optimal, beygelzimer2015online, agarwal2019boosting, jung2017online, jung2018online,brukhim2020online2}. 
In each of these works, a certain component prevents the result from applying to the full OCO setting. Either the weak learner has a parameter $\gamma = 1$, the convex decision sets are restricted, or the loss functions are restricted. 

Our work is the first to generalize online boosting to real-valued prediction in the full setting of online convex optimization, that allows for arbitrary convex decision sets and arbitrary approximation factor $\gamma$. This requires the use of a new extension operator, defined henceforth.

For extensive background on online learning and online convex optimization see  \cite{hazan2019introduction,shalev2012online}. 

The contextual experts and bandits problems have been proposed in \cite{langford2008epoch} as a decision making framework with large number of policies. In the online setting, several works study the problem with emphasis on efficient algorithms given access to an ERM oracle \cite{rakhlin2016bistro,syrgkanis2016improved,syrgkanis2016efficient,rakhlin2016bistro}.

Another related line of work in online learning studies the possibility of $\alpha$-regret minimization given access to approximation oracles with an approximate weak learner \cite{kakade2009playing, hazan2018online}. This weaker notion of $\alpha$-regret only generates approximately optimal decisions. In contrast, our result gives a stronger guarantee of standard regret, and the regret is compared to a stronger benchmark of convex combination of experts.

\section{Improper learning and the Extension operator}

\subsection{Preliminaries}

For a convex set $\K$ and scalar $c > 0 $ we denote by $c \K$ the set of all scaled points, i.e. 
$$ c \K = \left\{ c\x \mbox{ s.t. }  \x \in \K \right\} . $$

The algorithmic primitives introduced next use a smoothing operator. Let $f$ be $G$-Lipschitz continuous. The smoothing operator may be defined via inf-convolution or Moreau-Yoshida regularization: 
\[ M_\delta[f] = \inf_{y\in\reals^d} \left\{f(y) + \frac{1}{2\delta} \|x-y\|^2 \right\}.\]
The following lemma elucidates well known properties of such an operator (see \cite{beck2017first}). 
\begin{lemma}\label{lem:smoothing2}
    The smoothed function $\fhat_\delta = M_\delta[f] $ satisfies: \begin{enumerate}
        \item $\hat{f}_\delta$ is $\frac{1}{\delta}$-smooth, and $G$-Lispchitz.
        \item $ |\hat{f} _\delta (\x) - f(\x) | \le \frac{\delta G^2}{2}$ for all $ \x \in \reals^d$.
    \end{enumerate}
\end{lemma}

\subsection{The extension operator} 
The main difficulty we face is that the weak learner only has a $\gamma$-approximate regret guarantee. To cope with this, the algorithm we describe henceforth scales the predictions returned by the weak learner by a factor of $\frac{1}{\gamma}$. This algorithm, a rescaled variant of the Frank-Wolfe method, guarantees competitiveness with the convex hull of the base class, if we could somehow play actions in $\frac{1}{\gamma}\K$, instead of $\K$. Since these actions do not belong to the decision set, they are infeasible.

This presents a significant challenge. Ideally, we would like that for every action in $\gamma^{-1}\K$, one could find an action in $\K$ of comparable (or lesser) loss (for all loss functions). It can be seen that some natural families of functions, i.e. linear functions, do not admit such an operation. For linear loss functions, for example, extrema always occur on the boundary of the set.

To remedy this, we modify the loss function being fed to the Frank-Wolfe algorithm in the first place. We call this modification an {\it extension} of the loss function outside $\K$. Define $\dist(\x,\K)$ the Euclidean distance to $\K$, and the Euclidean projection $\Pi_\K$  as
\begin{align*}
    \dist(\x,K) = \min_{\y \in \K} \|\y -\x\|, \\
    \Pi_\K(\x) = \argmin_{\y\in \K} \|\y-\x\|
\end{align*}
\begin{definition}[$(\K,\delta,\kappa)$-extension] \label{defn:extension}
The extension operator over $\K \subseteq \reals^d$ %, with respect to Bregman divergence $B_R$, 
is defined as:
$$  X_{\K,\delta,\kappa}[f ](x)  = M_\delta[  f(x)  + \kappa \cdot \dist(\x,\K)  ] . $$ 
\end{definition}
We henceforth denote by $G$ an upper bound on the norm of the gradients of the loss functions over $\K$, and set 
$ \kappa = G  = \max_t \max_{\x \in \reals^d}  \| \nabla f_t(\x) \| . $
The important take-away from these operators is the following lemma, whose importance is crucial in the OCO boosting algorithm  \ref{alg:ocoboost}. The extended loss functions are smooth, agree with the original loss on $\K$, and possess the property that projection of any action in $\R^d$ onto $\K$ does not increment the extended loss value. In short, it permits projections of infeasible points that are obtained from the weak learners to the feasible domain without an increase in cost.

\begin{lemma} \label{lem:extensionX}
The $(\K,\delta,\kappa)$-extension of a $G$-lipschitz function $\fhat = X_{K,\delta,\kappa}[f]$ satisfies the following:
\begin{enumerate}
    \item For every point $\x \in \K$, we have $ | \fhat(\x) - f(\x) | \leq \frac{\delta G^2}{2}$.
    \item The projection of a point $\x$ onto $\K$ does not increase the $(\K,\kappa,\delta)$-extended function value by more than $G^2\delta$.
    $$   \fhat \left( \Pi_\K(\x)\right) \leq \fhat(\x) + G^2\delta . $$
\end{enumerate}
\end{lemma}

\begin{proof}
Since $\dist(\x,\K) = 0$ for all $x \in \K$, the first claim follows immediately from the properties of the smoothing operator as stated in Lemma \ref{lem:smoothing2}. Denote $\x_\pi =  \Pi_\K(\x)  $.
    \begin{align*}
    \fhat(\x_\pi) - \fhat(\x) &\leq  f(\x_\pi) - f(\x) - \kappa \cdot \dist(\x,\K)  + G^2\delta \\
    &= f(\x_\pi) - f(\x) - \kappa \| \x - \x_\pi\| + G^2 \delta \\ &\leq   G \|\x - \x_\pi\| - \kappa \| \x - \x_\pi\| + G^2 \delta  = G^2\delta 
    \end{align*}
Here the first inequality follows from Lemma \ref{lem:smoothing2}.2, and the second follows from the lipschitz assumptions on $f$. The last equality results from the choice $\kappa=G$.
\end{proof}

\section{Algorithm and Main Theorems}

Algorithm \ref{alg:ocoboost} efficiently converts a weak online learning algorithm into an OCO algorithm with vanishing regret in a black-box manner. The idea is to apply the weak learning algorithm on linear functions that are gradients of the loss. The algorithm then recursively applies another weak learner on the gradients of the residual loss, and so forth.

\begin{algorithm}[h]
\caption{{\bf BoOCO} = Boosting Online Convex Opt.}
\label{alg:ocoboost}
\begin{algorithmic}[1]
\STATE Input: $N$ copies of the $\gamma$-WOCL $\W^1, \W^2, \ldots, \W^N$, parameters $(\eta_1, \dots \eta_N)$, $\delta$, $\kappa = G$.
\FOR{$t=1$ {\bfseries to} $T$}
    \STATE Receive context $c_t$.
    \STATE Choose $\x_t^0\in\K$ arbitrarily.
    \FOR{$i=1$ {\bfseries to} $N$}
    	\STATE Define $\x_t^i = (1 - \eta_i) \x_t^{i-1} + \eta_i \frac{1}{\gamma} \W^i(c_t)$.
    \ENDFOR
    \STATE Predict $\x_t = \Pi_\K(\x_t^N )$ and suffer loss $f_t(\x_t)$.
    \STATE Obtain loss function $f_t$, create $\fhat_t = X_{\K,\delta,\kappa}[f_t]$.
    \FOR{$i=1$ {\bfseries to} $N$}
        \STATE Pass to $\W^i$ the linear loss function $f_t^i$.
        $$f_t^i(\x) = \nabla \fhat_t(\x_t^{i-1}) \cdot \x  . $$
    \ENDFOR
\ENDFOR
\end{algorithmic}
\end{algorithm}

The main performance guarantee we prove is summarized in the following theorem.

\begin{theorem}[Main] \label{thm:oco-boost-convex}
The actions $\x_t$ generated by Algorithm~\ref{alg:ocoboost} with $\delta = \sqrt{ \frac{ D^2}{\gamma N}  }, \eta_i = \min \{\frac{2}{i}, 1\}$ satisfy
\begin{align*}
\sum_{t=1}^T f_t(\x_t)\ - \min_{h \in CH(\H)} \sum_{t=1}^T f_t( h(c_t) ) \leq   \frac{ 4G  D T}{\gamma\sqrt{  N}}  +  \frac{2GD}{\gamma } \mbox{Regret}_T(\W)  .
\end{align*}
\end{theorem}

\section{Analysis}
Define $\fhat_t = X[f_t] = M_\delta [ f_t  +  G\cdot \dist(\x,\K) ]$. We apply the setting of $\kappa=G$, as required by Lemma \ref{lem:extensionX}, and by Lemma \ref{lem:smoothing2}, $\fhat_t$ is $\frac{1}{\delta}$-smooth.

Let $h^* = \argmin_{h\in CH(\H)}\sum_{t=1}^T f_t(h(c_t))$ be the best hypothesis in the convex hull in hindsight. We define $\x_t^* = h^*(c_t)$ as the decisions of this hypothesis.

The main crux of the proof is given by the following lemma.
\begin{lemma} \label{lem:main-analysis}
Suppose $\fhat_t$'s are ${\beta}$-smooth, and $\hat{G}$ lipschitz. Then,
\begin{eqnarray*}
\sum_{t=1}^T \fhat_t(\x_t^N)\ -  \sum_{t=1}^T \fhat_t( \x^\star_t )  \leq  \frac{2 {\beta} D^2 T}{\gamma^2 N} +  \frac{\hat{G}D}{\gamma} \regret_T(\W) .
\end{eqnarray*}
\end{lemma}
\begin{proof}
Define for all $i = 0, 1, 2, \ldots, N$, 
$$\Delta_i = \sum_{t=1}^T \left(\fhat_t(\x_t^i) - \fhat_t(\x^\star_t)\right) . $$ 
Recall that $\fhat_t$ is ${\beta}$ smooth by our assumption. Therefore:
	\begin{align*}
	  \Delta_i  = & \sum_{t=1}^T \left[ \fhat_t(\x_t^{i-1} + \eta_i ( \frac{1}{\gamma} \W^i(c_t) - \x_t^{i-1})) - \fhat_t(\x^\star_t) \right]\\
	\leq & \sum_{t=1}^T \left[ \fhat_t(\x_t^{i-1}) - \fhat_t(\x^\star_t)  + \eta_i \nabla \fhat_t(\x_t^{i-1}) \cdot ( \frac{1}{\gamma} \W^i(c_t) - \x_t^{i-1})  + \frac{\eta_i^2{\beta}}{2} \| \frac{1}{\gamma} \W^i(c_t) - \x_t^{i-1} \|^2  \right] 
	\end{align*}
By using the definition and linearity of $f_t^i$, we have
\begin{align*}
    \Delta_i \leq&  \sum_{t=1}^T \left[\fhat_t(\x_t^{i-1}) - \fhat_t(\x^\star_t)  +\eta_i ( f_t^i( \frac{1}{\gamma} \W^i(c_t))  - f_t^i(\x_t^{i-1}))  + \frac{\eta_i^2{\beta} D^2}{2 \gamma^2} \right]  \\
	 =& \Delta_{i-1}  + \sum_{t=1}^T \eta_i ( \frac{1}{\gamma} f_t^i(  \W^i(c_t))  - f_t^i(\x_t^{i-1})) + \sum_{t=1}^T\frac{\eta_i^2{\beta} D^2}{2 \gamma^2} 
\end{align*}
Now, note the following equivalent restatement of the WOCL guarantee, which again utilizes linearity of $f_t^i$ to conclude: linear loss on a convex combination of a set is equal to the same convex combination of the linear loss applied to individual elements.
\begin{align*}
    \frac{1}{\gamma} \sum_{t=1}^T f_t^i (\W^i(c_t)) \leq &\min_{h\in \H} \sum_{t=1}^T f_t^i (h(c_t)) + \frac{\hat{G}D \textrm{Regret}_T(\W)}{\gamma}\\
    =& \min_{h\in CH(\H)} \sum_{t=1}^T f_t^i (h(c_t)) + \frac{\hat{G}D \textrm{Regret}_T(\W)}{\gamma}
\end{align*}
Using the above and that $h^*\in CH(\H)$, we have
\begin{align*}
     \Delta_i   & \leq  \Delta_{i-1}  + \sum_{t=1}^T [ \eta_i \nabla \fhat_t(\x_t^{i-1}) \cdot (\x_t^\star - \x_t^{i-1})   + \frac{\eta_i^2{\beta} D^2}{2 \gamma^2 }     ] + \eta_i \frac{\hat{G}D}{\gamma} \regret_T(\W ) \\
    & \leq  \Delta_{i-1}  (1 -  \eta_i )  + \frac{\eta_i^2{\beta} D^2 T }{2 \gamma^2 }  + \eta_i  {R_T} 
\end{align*}
where the last inequality uses the convexity of $\hat{f}_t$ and $R_T = \frac{\hat{G}D}{\gamma} \regret_T(\W )$.
We thus have the recurrence
$$ \Delta_i \leq \Delta_{i-1} (1 - \eta_i) +  \eta_i^2 \frac{{\beta} D^2 T }{2 \gamma^2 }  + \eta_i  {R_T} . $$
Denoting $\hat{\Delta}_i = \Delta_i - {R_T}  $,	we are left with
$$ \hat{\Delta}_i \leq \hat{\Delta}_{i-1} (1 - \eta_i) +  \eta_i^2 \frac{{\beta} D^2 T }{2 \gamma^2 }   . $$

\begin{lemma} \label{lemma:FW-recursion}
Let $\{ h_t \} $ be a sequence that satisfies the recurence 
$$ h_{t+1} \leq h_t (1 - \eta_t) + \eta_t^2 c . $$
Then taking $\eta_t = \min\{1,\frac{2}{t}\} $ implies
$$ h_t \leq \frac{4c}{t} . $$ 
\end{lemma}
\

This is a recursive relation that can be simplified by applying Lemma \ref{lemma:FW-recursion}, taken from \cite{hazan2019introduction}.  
. Specifically, we obtain that $\hat{\Delta}_N \leq \frac{2 {\beta} D^2 T}{\gamma^2 N}$.
\end{proof}

We are ready to prove the main guarantee of Algorithm \ref{alg:ocoboost}.

\begin{proof}[Proof of Theorem \ref{thm:oco-boost-convex}]
Using both parts of Lemma \ref{lem:extensionX} in succession, we have
\begin{align*}
 \sum_{t=1}^T f_t(\x_t)\ -  \sum_{t=1}^T f_t( \x^\star_t ) 
& \leq \sum_{t=1}^T \fhat_t(\x_t)\ - \sum_{t=1}^T \fhat_t( \x^\star_t ) + \delta G^2 T \\
& \leq  \sum_{t=1}^T \fhat_t(\x_t^N)\ -  \sum_{t=1}^T \fhat_t( \x^\star_t ) + 2 \delta G^2 T \label{eqn:shalom4}
\end{align*}
Next, recall by Lemma \ref{lem:smoothing2}, that $\fhat_t$ is $\frac{1}{\delta}$-smooth. By applying Lemma \ref{lem:main-analysis}, and optimizing $\delta$, we have \begin{align*}
 \sum_{t=1}^T f_t(\x_t)\ -  \sum_{t=1}^T f_t( \x^\star_t ) 
& \leq  2 \delta G^2 T +  \frac{2   D^2 T}{\delta \gamma^2 N} +  \frac{\hat{G}D}{\gamma} \regret_T(\W) \\
& =  \frac{ 4 G  D T}{\gamma \sqrt{N}}  +  \frac{\hat{G}D}{\gamma} \regret_T(\W) 
\end{align*}
\begin{lemma}\label{lem:dist}
$\dist(\x,\K)$ is $1$-Lipschitz,
\end{lemma}
Using the above lemma, we claim that $\hat{G} \leq 2G$. Therefore, using Lemma \ref{lem:smoothing2}, $\|\nabla f_t^i(x_t^i)\|=\|\nabla \hat{f}_t(x_t^i))\| \leq 2G$.
\end{proof}

\begin{proof}[Proof of Lemma~\ref{lem:dist}]
Observe the following.
\begin{align*}
 \dist (\x, \K) - \dist(\y, \K) 
& = \|\x-\Pi_\K(\x)\|  - \|\y-\Pi_\K(\y)\| \\
& \leq \|\x-\Pi_\K(\y)\|  - \|\y-\Pi_\K(\y)\| & \mbox{ $\Pi_\K(\y)\in\K$}\\
& \leq \|\x-\y\| & \mbox{ triangle inequality}
\end{align*}
\end{proof}

\begin{proof}[Proof of Theorem \ref{lemma:FW-recursion}]
We give a proof for completeness. This is proved by induction on $t$. 

{\bf Induction base.}  For $t=1$, we have 
$$ h_2 \leq h_1 (1-\eta_1 ) + \eta_1^2 c = c \leq 4c $$

{\bf Induction step.} 
\begin{align*}
 h_{t+1} & \leq (1- \eta_t) h_t + \eta_t^2 c  \\
 & \leq \left(1- \frac{2}{t} \right) \frac{4c}{ t} + \frac{4c}{t^2}  & \mbox{induction hypothesis}\\
 & = \frac{4c}{t} \left( 1 - \frac{1}{t} \right) \\
 & \leq  \frac{4c}{t} \cdot  \frac{t}{t+1} & \mbox{$\frac{t-1}{t} \leq \frac{t}{t+1} $ }   \\
 & = \frac{4c}{t+1} 
 \end{align*}
\end{proof}

\section{Boosting for Bandit Linear Optimization}

Recall that the setting of Bandit Linear Optimization (BLO) is exactly the same as OCO, but (1) the cost functions are linear, and (2) the feedback for the decision maker is the cost it incurs (no gradient information). We use the notation that our linear cost function at time $t$ is:
$$ f_t(\x) = f_t^\top \x \ ,  \ f_t \in \reals^d . $$ 
To boost in the linear bandit setting, we apply techniques for unbiased gradient estimation. Namely, we use randomness to create a random linear function whose expectation is the actual linear function. We can then hope to use the previous algorithm on the random linear functions and obtain a probabilistic regret bound. 

%Unfortunately, an unbiased estimate of gradient of the loss may not be enough to compute an unbiased estimate of the gradient on the extension, as the previous algorithm requires. We define yet another extension operator that utilizes randomized smoothing. Unlike Moreau-Yoshida smoothing, a randomized smoothing operator 

For the rest of this section, we assume the following,
\begin{assumption}
The convex decision set $\K$ contains the unit simplex $K \supseteq \Delta_d$.
\end{assumption}

This assumption is very unrestrictive: by scaling and rotation, any set $\K$ which is non-degenerate contains the simplex. Scaling changes regret by a constant factor, and thus we scale so as to contain the unit simplex for notational convenience. An axis-aligning rotation does not affect the regret bound. We use standard basis aligned simplex for notational simplicity.

We can now use standard randomized exploration to create an unbiased estimator of the loss function, as per the following scheme. Let $b_t$ be a Bernoulli random variable with parameter $\eta$. Let $i_t$ be chosen uniformly at random from $i_t \in \{1,2,...,d\}$.
\begin{eqnarray}\label{eqn:randomf_t}
\ftil_t(i) = \mythreecases{0} {$i \neq i_t$ } {0} {$i=i_t, b_t = 0$} {\frac{d}{\eta} \times f_t(i_t) } {$i=i_t, b_t=1$} 
\end{eqnarray}

Clearly the expectation of the random vector $\ftil_t$ equals $\E[\ftil_t] = f_t$. We can now use this random variable as feedback for Algorithm \ref{alg:ocoboost}, as given in Algorithm \ref{alg:boost-blo}.

\begin{algorithm}
\caption{{\bf BoBLO} = Boosting Bandit Linear Opt.}
\label{alg:boost-blo}
\begin{algorithmic}[1]
\STATE Input: parameters $\eta$, Algorithm \ref{alg:ocoboost} instance $\mA$.
\FOR{$t=1$ {\bfseries to} $T$}
    \STATE Observe context $c_t$.
    \STATE Draw random Bernoulli $b_t$ w. parameter $\eta$.
    \IF {$b_t = 1$ }
        \STATE Pick a coordinate basis $i_t\in \R^d$ vector randomly.
        \STATE Play $\x_t = i_t$.
        \STATE Pass $\ftil_t$ as per Equation~\eqref{eqn:randomf_t} to $\mA$.
    \ELSE
        \STATE Play $\x_{t} = \mA(c_t)$, and pass the $\mathbf{0}$ loss vector to $\mA$.
    \ENDIF
\ENDFOR
\end{algorithmic}
\end{algorithm}

Notice that Algorithm \ref{alg:boost-blo} calls upon weak learners for OCO with full information. For this algorithm, our main performance guarantee is as follows. The resulting bound is sub-linear whenever the full-information regret bound is sublinear. Therefore, boosting in bandit linear settings is feasible whenever the full-information version is feasible, albeit at a slower rate.

\begin{theorem}[Boosting for Bandit Linear Optimization]\label{thm:blo}   The predictions $\x_t$ generated by Algorithm~\ref{alg:boost-blo} satisfy
\begin{align*}
  \mathbb{E}\left[\sum_{t=1}^T f_t(\x_t)\right]\ - \inf_{h \in CH(\H)} \sum_{t=1}^T f_t( h(c_t) )  
 & \leq    GD \sqrt{ T \left(\frac{ 16 d T}{  \gamma \sqrt{N}}  +  \frac{8 d }{ \gamma} \regret_T\left(\W \right)\right)} 
\end{align*}   
\end{theorem}

\begin{proof} Let $\y_t=\mA(c_t)$. Observe that
\begin{align*}
 \E \left[\sum_{t=1}^T f_t(\x_t)\right]\ - \inf_{h \in CH(\H)} \sum_{t=1}^T f_t(  h(c_t) ) 
& \leq \E \left[\sum_{t=0}^T ( \widetilde{f}_t(\y_t)\ -   \widetilde{f}_t( \x^\star_t )) \right]+ \eta  GD T \notag \\
& \leq    \frac{ 4 D \tilde{G}  T}{ \gamma \sqrt{N}}  +  \frac{2D \tilde{G} }{\gamma^2 } \mbox{Regret}_{ T}(\W ) + \eta GDT  &  \mbox{Theorem \ref{thm:oco-boost-convex}}\\
& \leq    \frac{ 4 D dG  T}{ \eta\gamma \sqrt{N}}  +  \frac{2dD G }{\eta\gamma} \mbox{Regret}_{ T}(\W ) + \eta GD T  
\end{align*}
Above, we use $\tilde{G}$ for an upper bound on the gradients of $\ftil_t$. Lastly, by construction the gradients of the loss function $\ftil_t$ are bounded by $\frac{d G}{\eta}$. Balancing $\eta$ concludes the claim
\end{proof}

\subsection{Application to contextual bandits}
The contextual bandits problem is one of the most well studied special cases of contextual BLO. In multi-armed contextual bandits, the decision set is a set of $K$ discrete arms, the losses are $[0, 1]$-bounded, and the feedback available to learner is solely the loss of the arm it picks. This is exactly BLO for the decision set $\K = \Delta_K$. We state a corollary of our result for this setting.

\begin{corollary}[Multi-armed Contextual Bandits]\label{thm:cmab}   The predictions $\x_t$ generated by Algorithm~\ref{alg:boost-blo} with the decision set $\K=\Delta_K$ and the loss set $[0,1]^K$ satisfy
\begin{align*}
  \mathbb{E}\left[\sum_{t=1}^T f_t(\x_t)\right]\ - \inf_{h \in\H} \sum_{t=1}^T f_t( h(c_t) )  
 & \leq    K \sqrt{ T \left(\frac{ 16 T}{  \gamma \sqrt{N}}  +  \frac{8 }{ \gamma} \regret_T\left(\W \right)\right)} 
\end{align*}
\end{corollary}

\section{Boosting for Stochastic Contextual Optimization}
In this section we give an alternative viewpoint of our boosting results from the lens of stochastic contextual optimization. The mathematical development is analogous to the previous results, but the statement of results may be of independent interest in statistical learning theory and contextual learning. 

Let $\K\subseteq \reals^d$ be the (convex) decision set, and $\C$ be the set of possible contexts. In the Stochastic Contextual Optimization problem the aggregate loss attributable to a hypothesis is the expected cost of its actions with respect to a joint distribution $\D$ over the context set $\C$ and the set of convex cost functions over the decision set $\K$. 
$$F_\D(h) = \E_{(f , c) \sim \D } [f( h(c) ) ] . $$ 

Instead of explicit optimization over the hypothesis class, we assume that we have access to experts that are approximately competitive with the base hypothesis class. Formally, we define a weak optimizer as: 
\begin{definition} \label{def:SCO-WOLL}
A {\bf $\gamma$-weak contextual optimizer}  is a learning algorithm that, when given samples from a distribution $\D$ over unit linear loss functions and contexts, outputs a mapping $\W:\C\to\K$ such that 
\begin{equation*} 
F_\D(\W) \leq \gamma \cdot \min_{h \in \H} F_\D(h)  + \eps .
 \end{equation*}
\end{definition}
 %  \E_{f_t \sim \D}  f_t(\W(c))  \le \gamma \cdot \underset{h \in \H}{\min} \E_{f_t\sim \D}  f_t(h(c))  + \eps_T =

Typically $\eps$ scales as $\sim \frac{1}{\sqrt{m}} $ when the learner is given $m$ samples from  the distribution $\D$.

%The function $F$ can now be seen to have important properties w.r.t. the simplex $\CH(\H)$, namely: \eh{should we talk about Jacobians here?} 
%\begin{enumerate}
%\item
%The function $F$ is convex as a function of $h$. To see this, notice that by definition $F$ is an expectation over functions, and it suffices that for every single function $f_t$ is convex in $h$, which it is by definition. 

The results in this section do {\bf not} follow blackbox from the previous ones. While online boosting algorithms operate in more permissive settings than statistical ones, as they do not require the context and loss vectors to be sampled i.i.d., the corresponding assumption on the weak learners for the online setting are stronger too.
%\item
%If the functions $f_t$ are $\beta$-smooth as a function of $h$, then so is the function $F$. 

%\end{enumerate}

\subsection{Algorithm and main theorem}

Algorithm \ref{alg:scoboost} below makes use of $N$ weak optimiziers, and generates a hypothesis $h:\C\to\K$ such that the following theorem holds. Notice that the resulting guarantee delivers a solution which is stronger in two aspects than the weak optimization oracle: 
\begin{enumerate}
\item
The resulting solution competes with the convex hull of $\H$, rather than the best fixed $h \in \H$. %We defined a solution in the convex hull of $\H$, denoted $h \in \CH(\H)$, as 
%$$ F(h) = \E_{f,c\sim \D} [ f_t( \sum_h h_h h(c) ) ] .$$ 
\item
The resulting solution removes the $\gamma$ approximation factor of the weak optimizers.
\end{enumerate}

\begin{theorem}[Main-SCO] \label{thm:co-boost-convex}
The hypothesis $h$ generated by Algorithm~\ref{alg:scoboost} with parameter $\delta = \sqrt{ \frac{ D^2}{\gamma N}  }$ satisfy
$$ F_\mathcal{D}(h)   - \inf_{h^\star \in \CH(\H)} F_\mathcal{D}(h^\star)  \leq   \frac{ 4G  D }{\gamma\sqrt{  N}}  +  \frac{2GD}{\gamma } \eps . $$
\end{theorem}

%The proof of this theorem follows from applying a reduction from regret minimization to batch learning commonly known as ``online to batch". Specifically, apply Theorem ... from \cite{ocobook}, we have  
\begin{algorithm}[h]
\caption{{\bf B4CO} = Boosting for Contextual Opt.}
\label{alg:scoboost}
\begin{algorithmic}[1]
\STATE Input: $\gamma$-weak contextual optimizer $\O$, parameters  $\{\eta_1,\dots \eta_N\}$, $\delta$, $\kappa = G$, distribution $\D$.
\STATE Choose $h^0$ arbitrarily from $\H$.
\FOR{$i=1$ {\bfseries to} $N$}
%\STATE Sample $T$ examples and losses $\{f_t, c\}$.
\STATE Define a distribution $\D^i$ specified via the following sampling oracle:\\
		\hspace{1cm} Sample $(f,c)\sim \D$, and define $\fhat = X_{\K,\delta,\kappa}[f]$\\
		\hspace{1cm} Define a linear loss function $$f^i(\x) = \nabla \fhat(h^{i-1}(c))\cdot \x$$\\
		\hspace{1cm} Output $(f^i, c)$.
\STATE Call a weak optimizer on the distribution $\D^i$ to get a hypothesis $\W^i$. 

    \STATE Define $h^i = (1-\eta_i) h^{i-1} +  \frac{\eta_i}{\gamma} \W^i $ %, and $\x_t = \Pi_\K(\A^N(c)  )$. 
        
\ENDFOR
\STATE Return predictor $h$, defined as  $h(c) = \Pi_\K( h^N(c) ) $.
\end{algorithmic}
\end{algorithm}

As before, for a convex loss function $f$, define $\fhat = X_{\K,\delta,G}[f]$, and note that $\fhat_t$ is $\frac{1}{\delta}$-smooth. Notations we use below:
\begin{enumerate}
\item  $F(h) = \E_{(f,c) \sim \D} [f ( h (c)) ]  $
    \item  $\hat{F}(h) = \E_{(f,c) \sim \D} [\fhat ( h(c)) ] $
\end{enumerate}

\begin{lemma} \label{lem:main-analysis2}
Suppose, for any $f\in \F$, $\fhat$ is ${\beta}$-smooth, and $\max_{f\in \F, x\in \K}\| \nabla \fhat(\x) \| \leq \hat{G}$. Then,
\begin{eqnarray*}
\hat{F}(h) \ -  \hat{F}(h^\star  )  \leq  \frac{2 {\beta} D^2 }{\gamma^2 N} +  \frac{\hat{G}D}{\gamma} \eps .
\end{eqnarray*}
\end{lemma}
\begin{proof}
Let $h^*=\argmin_{h\in CH(\H)} F(h)$. Denote for all $i = 0, 1, 2, \ldots, N$, 
$\Delta_i = \hat{F}(h^i) - \hat{F}(h^\star)$.

Recall that $\fhat_t$ is ${\beta}$ smooth by our assumption. Thus, 
\begin{align*}
 \Delta_i 
=&\   \hat{F}(h^i) - \hat{F}(h^\star) \\
= & \  \E_{f,c \sim \D}  [\fhat(h^i(c))]  - \hat{F}(h^\star)  \\
= &\  \E_{f,c \sim \D}  \left [\fhat(h^{i-1}(c)  + \eta_i ( \frac{1}{\gamma} \W^{i} (c)- h^{i-1}(c) )  \right] - F(h^\star)\\
\leq &\ \E_{f,c \sim \D} [ \fhat_t(h^{i-1}(c))   + \eta_i \nabla \fhat(h^{i-1}(c)) \cdot ( \frac{1}{\gamma}  \W^{i}(c) - h^{i-1}(c))  + \frac{\eta_i^2{\beta}}{2} \| \frac{1}{\gamma} h^{i-1}(c) - \W^{i}(c) \|^2  ] - \hat{F}(h^\star) \\
\leq & \E_{f,c \sim \D}  [  \fhat(h^{i-1}(c))  + \eta_i ( f^i( \frac{1}{\gamma} \W^i(c))  - f_t^i(h^{i-1}(c) ))+ \frac{\eta_i^2{\beta} D^2}{2 \gamma^2} ]  - \hat{F}(h^\star) 
\end{align*}
Using linearity of $f^i$, which implies that optimal aggregate loss values are equal over $\H$ and $CH(\H)$ hypotheses classes, and the definition of weak learning, we have 
\begin{align*}
	 \Delta_i &\leq  \Delta_{i-1} 
     +  \eta_i \E_{f,c \sim \D}  \left[ \nabla \fhat_t(h^{i-1}(c)) \cdot (h^\star(c) - h^{i-1}(c) )       \right]  + \frac{\eta_i^2{\beta} D^2}{2 \gamma^2 } + \eta_i \frac{\hat{G}D\eps}{\gamma} \\
    & \leq  \Delta_{i-1}  (1 -  \eta_i )  + \frac{\eta_i^2{\beta} D^2 }{2 \gamma^2 }  + \eta_i  \frac{\hat{G}D\eps}{\gamma}
    \end{align*}
where the last step follows from the convexity of $f$. We thus have the recurrence
$$ \Delta_i \leq \Delta_{i-1} (1 - \eta_i) +  \eta_i^2 \frac{{\beta} D^2  }{2 \gamma^2 }  + \eta_i  \frac{\hat{G}D\eps}{\gamma} . $$
As before, substituting $\hat{\Delta}_i = \Delta_i - \frac{\hat{G}D\eps}{\gamma}  $, and applying Lemma \ref{lemma:FW-recursion}, gives us $\hat{\Delta}_N \leq \frac{2 {\beta} D^2}{\gamma^2 N}$.
\end{proof}

\begin{proof}[Proof of Theorem \ref{thm:co-boost-convex}]
Using Lemma \ref{lem:extensionX}, we have,
\begin{align*}
F(h)\ -  F ( h^\star ) 
& = \E_{f,c} [ f(h(c)) ] \ -  \E_{f,c} [ f( h^\star(c) )] \notag \\
& \leq \E_{f,c} [ \fhat(h(c)) ] \ -  \E_{f,c} [ \fhat( h^\star(c) )] + \delta G^2  \\
& \leq  \E_{f,c} [ \fhat(h^N(c)) ] \ -  \E_{f,c} [ \fhat( h^\star(c) )] + 2 \delta G^2 \\
\end{align*}
By Lemma \ref{lem:smoothing2}, that $\hat{f}$ is $\frac{1}{\delta}$-smooth. By applying Lemma \ref{lem:main-analysis2}: \begin{align*}
& F(\mA)\ -  F ( h^\star )  \leq  2 \delta G^2  +  \frac{2   D^2 }{\delta \gamma^2 N} +  \frac{\hat{G}D}{\gamma} \eps 
%& =  \frac{ 4 G  D }{\gamma \sqrt{N}}  +  \frac{\hat{G}D}{\gamma} \eps
\end{align*}
Because $\dist(x,K)$ is $1$-Lipschitz, using Lemma \ref{lem:smoothing2}, $\|\nabla f_t^i(x_t^i)\|=\|\nabla \hat{f}_t(x_t^i))\| \leq 2G$. Choosing the suggested value of $\delta$ yields the proof.
\end{proof}

\section{Experimental Results}
    \begin{figure*}[]
        \centering
        \begin{tabular}{c|c|c|c|c|c|c}
        \hline\hline
        \multicolumn{7}{c}{\bf Boston dataset}\\
            \hline
            & WL & N=2 & N=3 & N=4 & N=5 & Improvement\\ \hline
            Decision Stumps & 1.000 & 0.941 & 0.891 & 0.853 & {\bf 0.821} & {\bf 17.9\%}\\ \hline Ridge Regression & 1.000 & {\bf 0.980} & 0.985 & 1.009 & 1.060 & {\bf 2.0\%} \\ \hline
            Tiny MLP & 1.000 & 0.987 & 0.966 & 0.934 & {\bf 0.920} & {\bf 8.0\%}\\
            \hline \hline
            \multicolumn{7}{c}{\bf Diabetes dataset}\\
            \hline
            & WL & N=2 & N=3 & N=4 & N=5 & Improvement\\ \hline
            Decision Stumps & 1.000 & 0.975 & 0.960 & 0.952 & {\bf 0.941} & {\bf 15.9\%} \\ \hline Ridge Regression & 1.000 & 0.986 & 0.974 & 0.965 & {\bf 0.956} & {\bf 4.4\%}\\ \hline
            Tiny MLP & 1.000 & 0.982 & 0.981 & 0.987 & {\bf 0.965} & {\bf 3.5\%} \\
            \hline\hline
            \multicolumn{7}{c}{\bf California Housing dataset}\\
            \hline
            & WL & N=2 & N=3 & N=4 & N=5 & Improvement\\ \hline
            Decision Stumps & 1.000 & 0.969 & 0.946 & 0.932 & {\bf 0.922} & {\bf 7.8\%} \\ \hline Ridge Regression & {\bf 1.000} & 1.019 & 1.041 & 1.066 & 1.094 & {\bf -1.9\%}\\  \hline
            Tiny MLP & 1.000 & 0.860 & 0.861 & 0.888 & {\bf 0.835} & {\bf 16.5\%} \\
            \hline \hline
        \end{tabular}
        \caption{Performance of the boosting algorithm on 3 different datasets, averaged across 20 runs, as a function of the weak learning class and the number of weak learners. The entries indicate normalized loss value with the base/weak learner loss set to one.}
        \label{tab:my_label}
    \end{figure*}

While the primary contribution of this work is theoretical, empirically testing our proposal serves as a sanity check for the theoretical results. Also, as stated in the related work, the proposed algorithm is the first of its kind in terms of being able to simultaneously operate with general convex losses and sets, and with multiplicatively approximate (inexact) weak learners. Consequently, the role of experiments here is confirmatory, and is not aimed at achieving state-of-the-art on the considered datasets.

To validate our results, we check if the proposed algorithm is indeed capable of boosting the accuracy of concrete instantiations of weak learners. Three online weak learners were considered: decision stumps, ridge regression, and a tiny multi-layer perceptron with one hidden unit trained via online gradient descent. The implementation for each was suitably adapted from Scikit-Learn \cite{scikit-learn}.

For each choice of weak learner class, we considered the performance of the boosting algorithm (Algorithm \ref{alg:ocoboost}) across multiple rounds of boosting or number of weak learners; the computational burden of the algorithm scales linearly with the latter. We evaluated the weak learners and the boosting algorithm on 3 publicly available datasets\footnote{The California Hosuing dataset \cite{pace1997sparse} and the Boston dataset \cite{harrison1978hedonic} can be found at \url{http://lib.stat.cmu.edu/datasets/}. The Diabetes dataset used here is present in UCI ML Repository (\url{https://archive.ics.uci.edu/ml/datasets/Diabetes}).} for regression with square loss, averaging each across 20 runs. The step size $\eta$ was set to $0.01$, and $\gamma$ was chosen to be $0.1$ -- these values were not tuned. We present the results in Table \ref{tab:my_label}. Further experimental details and code are detailed in the supplement.

The results demonstrate the boosting for online convex optimization does indeed succeed in boosting the accuracy while using only few weak learners (equivalently, within a few rounds of boosting), and therefore, with a reasonably small increase in computation. The improvement is most pronounced for decision stumps, since this is an especially impoverished base class. Also, note that a convex combination of linear models is linear. Therefore, boosting a linear model (like ridge regression) has no expressivity benefit. Mirroring this, the improvements in the ridge regression setting are least prominent, and in some cases negative.

\section{Conclusions}

We study the problem of online convex optimization in the contextual setting. To cope with the computational challenges, we take a boosting approach, and generalize the methodology of gradient boosting and online boosting to contextual online convex optimization. 

Our derivation introduces a new extension operator for convex functions over convex domains which may be of independent interest. 

\bibliographystyle{alpha}
\bibliography{main}

\end{document}